\documentclass[runningheads]{llncs}
\usepackage{makeidx}
\usepackage{graphicx}
\usepackage{amsmath,amssymb} 
\usepackage{color}
\usepackage{pifont}

\begin{document}
\pagestyle{headings}
\mainmatter
\def\GCPR16SubNumber{85}
\title{Identifying individual facial expressions by deconstructing a neural network}
\titlerunning{Identifying individual facial expressions by deconstructing a neural network}
\authorrunning{Arbabzadah et al.}
\author{Farhad Arbabzadah$^1$, Gr{\'e}goire Montavon$^1$,\\ Klaus-Robert M{\"u}ller$^{1,2}$, and Wojciech Samek$^3$}
\institute{$^1$ Machine Learning Group, Technische Universit\"at Berlin, Berlin, Germany\\
$^2$ Department of Brain and Cognitive Engineering, Korea University, Seoul, Korea\\
$^3$ Machine Learning Group, Fraunhofer Heinrich Hertz Institute, Berlin, Germany\footnote{Paper will appear in the Proceedings of the 38th German Conference on Pattern Recognition (GCPR 2016).}}
\maketitle

\begin{abstract}
This paper focuses on the problem of explaining predictions of psychological attributes such as attractiveness, happiness, confidence and intelligence from face photographs using deep neural networks. Since psychological attribute datasets typically suffer from small sample sizes, we apply transfer learning with two base models to avoid overfitting. These models were trained on an age and gender prediction task, respectively. Using a novel explanation method we extract heatmaps that highlight the parts of the image most responsible for the prediction. We further observe that the explanation method provides important insights into the nature of features of the base model, which allow one to assess the aptitude of the base model for a given transfer learning task. Finally, we observe that the multiclass model is more feature rich than its binary counterpart. The experimental evaluation is performed on the 2222 images from the 10k US faces dataset containing psychological attribute labels as well as on a subset of KDEF images.
\end{abstract}
\section{Introduction}
Deep Convolutional Networks (ConvNets) \cite{le1990handwritten} are the architecture of choice for many practical problems such as large-scale image recognition \cite{krizhevsky2012imagenet,he2015deep}. Typically, these networks have millions of free parameters that are learned from large datasets. The great success of ConvNets and an abundance of hardware and software frameworks \cite{jia2014caffe} has led to the availability of a set of pre-trained models for specified tasks. Retraining these pre-trained models to a new task allows to successfully use ConvNets also in small sample settings \cite{yosinski2014transferable,donahue2013decaf}. This procedure is an instance of the transfer learning paradigm \cite{caruana1997multitask}. In a nutshell, a model is trained on a domain related to the problem at hand, then adapted to the target domain with only few data points.

In this contribution a pre-trained deep neural network trained for gender and age classification \cite{levi2015age} is used and subsequently retrained to predict psychological attributes such as attractiveness, happiness, confidence and intelligence from face photographs. Since most labelled attribute datasets are rather small, we use a pre-trained model and thus harvest from the rich representation that the neural network has learned on a related problem. If the tasks are sufficiently similar, we can transfer the knowledge acquired by the pre-trained model from one domain, e.g., age or gender, to another, e.g., happiness or attractiveness.
Our study is based on the 10k US faces dataset \cite{bainbridge2013intrinsic}. Note that only a subset of 2222 images in that dataset are labelled with the needed attributes. Therefore we retrain the neural network \cite{levi2015age} to reproduce the human assessment labels from the Bainbridge dataset. 
However, our focus is not primarily the excellent prediction of these attributes as such, but the explanation thereof. So, we focus on the question of {\it what} makes the neural network assign certain attractiveness, happiness scores etc.\ to a novel test image and whether this assignment corresponds to what we as humans expect.
For this we apply a recently proposed explanation method termed Layer-wise Relevance Propagation (LRP) \cite{bach2015pixel} that allows a better understanding of what a neural network is using as decisive features on a single sample basis. We visualize the rationale of the trained predictors using this method, study its robustness and relate the results to human expectations. 
\section{Understanding Neural Networks}
\begin{figure}[t]
\centering
\includegraphics[height=5.5cm]{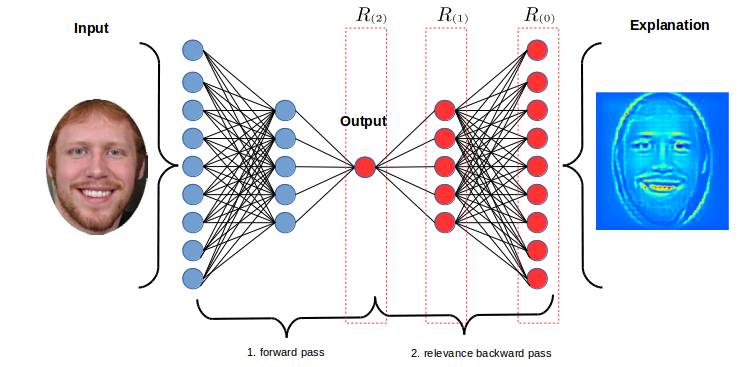}
\caption{
Illustration of the Layer-wise Relevance Propagation (LRP) for happiness classification. The first step of the procedure is a regular forward pass (blue nodes) that yields the classification score $f(x) = R_{(2)}$, the first red node. The second step is a backward pass (red nodes) that passes through the network in a reverse order. In the backward pass the score $R_{(2)}$ is redistributed layer by layer (indicated by the red rectangles). Each node at the lower layer will receive a relevance score relative to the activations $ a_{j}$ of each node based on one of the rules in eq.~(\ref{eq:firstrule}) or  (\ref{eq:secondrule}). 
}
\label{fig:example}
\end{figure}

The explanation method we are going to use in the following experiments is Layer-wise Relevance Propagation (LRP) \cite{bach2015pixel}. The basic idea is to perform a backward pass of the final score for a given image $x$ through the neural network, such that one finds a mapping between the score and the images input pixels, which reflects the {\it relevance} $R_p$ of the respective pixels $p$ of the image  $x$ for the network's decision. The first proposition of the LRP method is the conservation of relevance between layers. Let $(l)$ and $(l+1)$ be adjacent layers, then the conservation of relevance is expressed as:
\begin{align}
   \sum_{i} R_{i}^{(l)} = \sum_{j} R_{j}^{(l + 1)},
\end{align}
where the sums run over all neurons in respective layers. The conservation principle ensures that the network's output $f(x)$ is redistributed to the input domain such that $f(x) = \sum_p R_p$.
The relative distribution of the relevance scores on the input domain reflects the importance of the respective pixels for the overall score.
We call the image of the pixel-wise relevance scores a {\it heatmap}, which can be visualized and interpreted (see Fig.\ \ref{fig:example}).

The individual assignment of relevance scores can be done using two methods. Let  $z_{ij} = a_{i}^{(l)}w_{ij}^{(l,l+1)}$
 be the weighted activation of neuron $i$ onto neuron $j$ in the following layer. 
The first rule assigns a weighted average of weighted activations $z_{ij}$ to each node. Mathematically, this rule can be written as
\begin{align}
  R_{i}^{(l)} =  \sum_{j} \frac{
  	z_{ij}
  }
  {
  \sum_{i^{\prime}} z_{i^{\prime}j} + \epsilon \cdot \mathrm{sign}(\sum_{i^{\prime}} z_{i^{\prime}j})
  } R_{j}^{(l + 1)}.
  \label{eq:firstrule}
\end{align}
The parameter $\epsilon$ in the first rule is only relevant if the denominator in the above expression is zero. Then setting $\epsilon>0$ will ensure, that there is no division by zero. Note that by setting $\epsilon\neq0$ the conservation of relevance between layers will no longer hold. In this case one can add an additional renormalization step before the next LRP-pass, which will ensure that the sum of relevances is preserved.

The second rule is based on the weighted average rationale as well. However, in the second rule the positive and negative weighted activations $z_{ij}^{+}$ and $z_{ij}^{-}$ are treated separately and weighted by two additional parameters $\alpha$ and $\beta$, which force a certain ratio between positive and negative contributions.
\begin{align}
  \sum_{i} R_{i}^{(l)} = \sum_{j} \big(
   \alpha\cdot \frac{z^{+}_{ij}}{\sum_{i^{\prime}}z_{i^{\prime}j}^{+}} + 
   \beta \cdot \frac{z^{-}_{ij}}{\sum_{i^{\prime}}z_{i^{\prime}j}^{-}}   
  \big) R_{j}^{(l + 1)}.
    \label{eq:secondrule}
\end{align}

Here, $z_{ij}^{+}$ and $z_{ij}^{-}$ denote the positive and the negative part of $z_{ij}$ respectively, such that $z_{ij}^{+} + z_{ij}^{-}=z_{ij}$. For $\alpha+\beta=1$ the relevance conservation property holds.

An empirical comparison of different explanation methods can be found in \cite{samek2015evaluating}.
For a more theoretical view on LRP we refer the reader to \cite{MonArXiv15} and for applications beyond image classification to \cite{ArrACL16,StuArXiv16}.
An implementation of LRP can be found in \cite{LapJMLR16} and downloaded from \texttt{www.heatmapping.org}.
The following heatmaps were obtained using eq.~(\ref{eq:secondrule}) with $\alpha=2$ and $\beta=-1$.

\section{Experimental Evaluation}
We retrain the models for age and gender by \cite{levi2015age} on the 10k US faces dataset \cite{bainbridge2013intrinsic} to predict individual ratings of attractiveness, happiness, confidence and intelligence for the 2222 images with these scores in the dataset. The retraining happens for each base model (age, gender) in two modes. In the first mode only the dense layers of the network are retrained for the new target scores. In the second mode the entire network including the convolution units are retrained. The presented subject faces are only those, that are part of the subset of 48 faces of the Bainbridge dataset \cite{bainbridge2013intrinsic}, which are licensed for publishing.

\subsection{Preprocessing \& Setup}
The individual image rating scores (1 to 9) are rescaled to the output domain $\left[0,\ 1\right]$. The input images are rescaled to fit the input format of the network (227x227 pixels and 3 color channels). A labelled subset of 2222 images of the 10k US faces dataset is used to train the different networks. The 2222 images are randomly split into training and test set, each with 1111 samples. In total, we use 16 different networks. Half of them with the age model as a base model and the other half with the gender model as the base model. Then each of the networks is either completely retrained or only the dense layers are adapted.

The images are composed of a white frame that contains an oval central region with the face. 
We train the model using stochastic gradient descent with a learning rate of 0.001 using a Nesterov momentum of 0.9 \cite{lecun2012efficient}. The weights are initialized for each of the base models to those	 provided by Levi \cite{levi2015age}. 
The annotated dataset is split equally into a test and training set.


\subsection{Performance and Analysis of Base Models}

\begin{figure}[!htb]
\centering
\includegraphics[height=3.5cm]{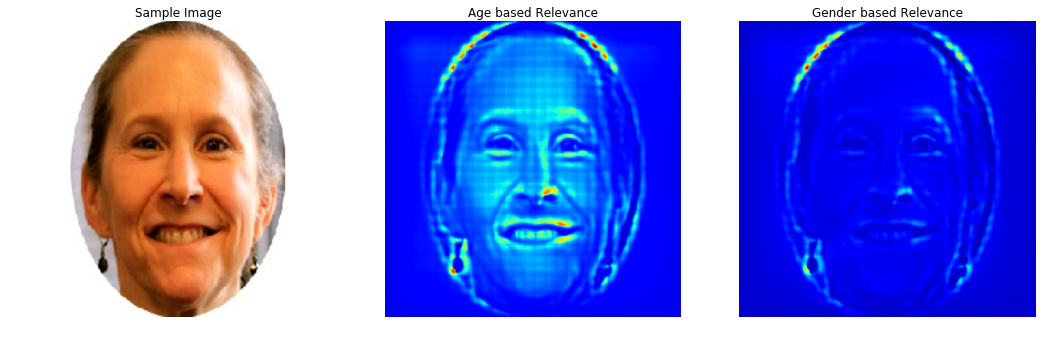}
\caption{
Left: Raw input image. Middle: Heatmap for a test-subject for the gender based model. Right: Heatmap for the age based model. 
}
\label{fig:base_model}
\end{figure}

\begin{table}[t]
\centering
\caption{Overview of mean absolute error of the 16 models.}
\begin{tabular}{lp{0.2cm}cp{0.2cm}cp{0.2cm}cp{0.2cm}cp{0.2cm}c}
Base       & & Retraining      & & Attractiveness & & Happiness & & Confidence & & Intelligence \\
\hline
Age & & Dense Only & & 0.43 & & 0.41 & & 0.37 & & 0.33  \\
    & & Full & & 0.42 & & 0.42 & & 0.34 & & 0.31  \\
\hline
Gender & & Dense Only & & 0.47 & & 0.36 & & 0.34 & & 0.35  \\
    & & Full  & & 0.43 & & 0.38 & & 0.52 & & 0.29 \\ 
\end{tabular}
\label{tab:mae}
\end{table}

Table \ref{tab:mae} lists the mean absolute error (MAE) for the prediction of scores versus the average score assigned by the human test group in the original dataset \cite{bainbridge2013intrinsic}. The error is given in terms of the human scaling from 1 to 9 points.
Interestingly, the gain of retraining the full model for the attractiveness assessment, does not yield a high improvement on the MAE, while the computational cost of retraining a full ConvNet are significantly higher than solely retraining the dense layers.

In transfer learning we want to select a base model that already captures as many as possible relevant features for the new task \cite{pan2010survey}. In case of our two base models, we would need to make a choice, whether to take the gender base model or the age base model. 
In the following we demonstrate that explanation methods, such as the LRP algorithm, are helpful in making an educated guess about which model to use as the basis for the transfer learning task.

Figure \ref{fig:base_model} illustrates for one sample image the difference of features that are picked up by the two models, if we use the same scale. It turns out that the age model picks up a lot more facial details, while the gender model is more sensitive to the overall shape of the face. The edges of the face seem to provide already quite a lot of information about the subject's gender. Note that the gender model is a binary classifier (female or male), while the age model is a multi-class classifier placing the subject into one of many possible age groups.
The multi-class structure of the age model might be the reason for it's feature richness. Wrinkles, for example, are only relevant for higher age groups, while teeth are universally relevant.

\subsection{Rate of Convergence}

\begin{figure}[!htb]
\centering
\includegraphics[height=9cm]{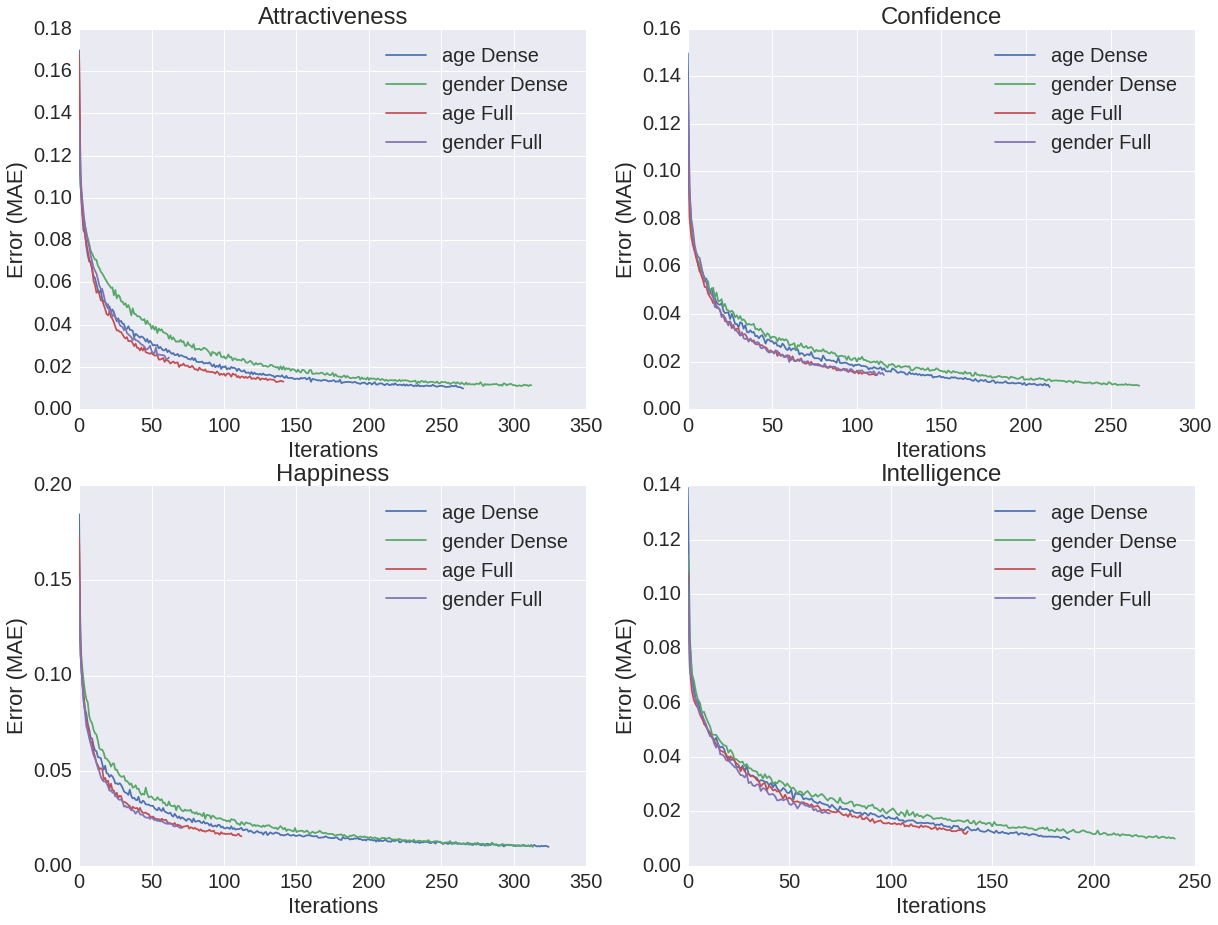}
\caption{
Training errors for attractiveness, happiness, confidence and intelligence ratings. The four curves correspond to the two base models and the two training modes (dense layers or full model retraining).
}
\label{fig:learning_curve}
\end{figure}

The learning curves in figure \ref{fig:learning_curve} show that the dense retrained models based on gender are learning slower than the models based on age. Thus, the above interpretation of the LRP heatmaps in terms of feature richness of the model correlates with the slower rate of convergence of the training error. Therefore, we conclude that the age based model captures more of those relevant features and thus the network based on the age model learns faster, both if we retrain the whole model or just the fully connected layers, as we can see from figure \ref{fig:learning_curve}.

This confirms that we can use the LRP method for the assessment of the aptitude of a base model for a given transfer learning task, especially if we possess additional knowledge about the structure of the problem (e.g., set of psychologically relevant features). If a model assigns the bulk of relevance to this significant feature subset, then we conjecture that the network applies a more {\it human-like} strategy as opposed to a model that is responding random artifact which happens to correlate with the task.

\subsection{Explanation for Happiness}
The 10k US faces dataset does not contain the same face in a different emotional state. Therefore, we have taken another dataset, the Karolinska directed emotional faces (KDEF) \cite{goeleven2008karolinska}, which contains the same adult person with different emotional states for the following analysis. We applied the same preprocessing as described above.
\begin{figure}[t]
\centering
\includegraphics[height=6cm]{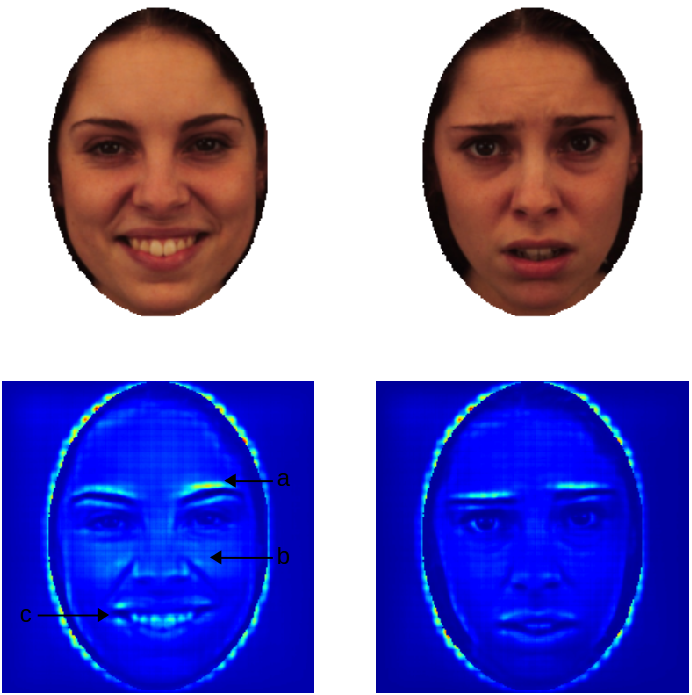}
\includegraphics[height=6cm]{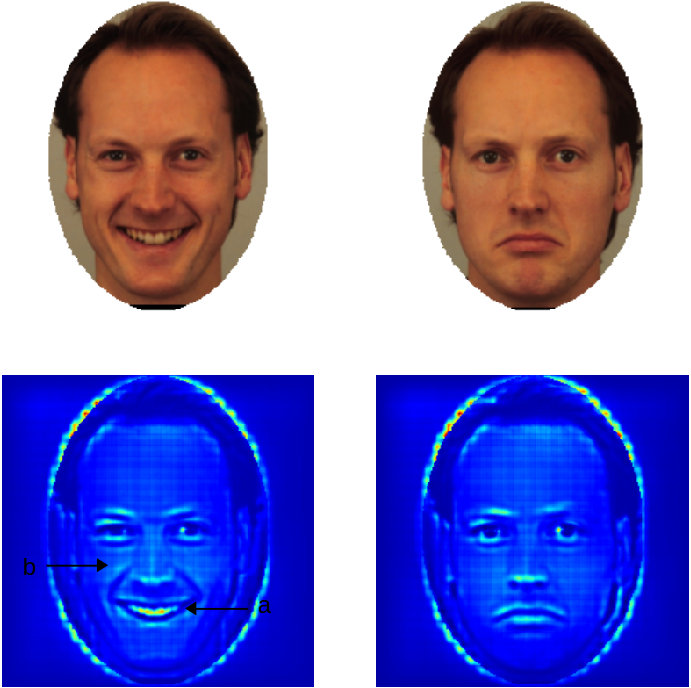}
\caption{
A female and male subject in a happy (left) and sad (right) state from the KDEF dataset. We see that the teeth and the smile as well as the dimples on the cheeks and the shape of the eyebrows are relevant.
}
\label{fig:cube_f}
\end{figure}

Figure \ref{fig:cube_f} depicts the heatmaps of a female and a male subject  in a happy and an unhappy state.
The heatmaps of the female subject have three principal regions of interest. Number one, pointed at by indicator arrow (c), is the mouth. Here we see that for the happy face the smile and the teeth are important. While the heatmap for the unhappy face also shows the teeth, they are not as relevant as in the happy case. Region number two, see indicator arrow (b), are the nostrils and the cheeks. The nostrils are wider compared to the sad state. Furthermore, the cheeks are lifted up forming dimples. Lastly, the outline of the eyebrows, as shown by indicator arrow (a), is marked relevant for both, the happy and unhappy face. However, the roundish shape of the outline for the happy face has more relevance assigned to it than the straight counterpart of the sad face.
For the male counterpart we make similar observations. Firstly, the shape of the lips seems to matter. Secondly, a smile showing the subject's teeth indicates happiness. Thirdly, the dimples matter too. In these examples most relevance is assigned to gender unspecific features such as smiles and dimples, however, some relevance is also assigned to the outline of the eyebrows which can be gender specific.

\subsection{Explanation for Age}
\begin{figure}[!htb]
\centering
\includegraphics[width=6cm]{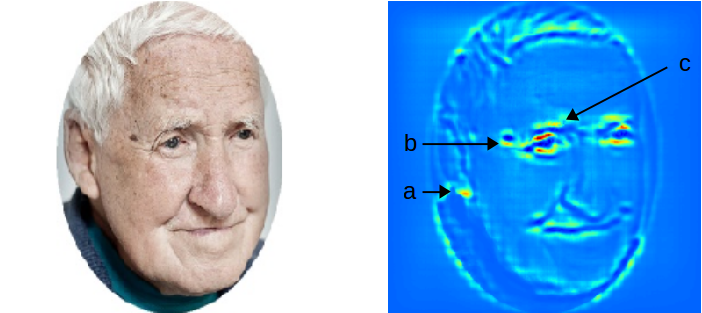}
\includegraphics[width=6cm]{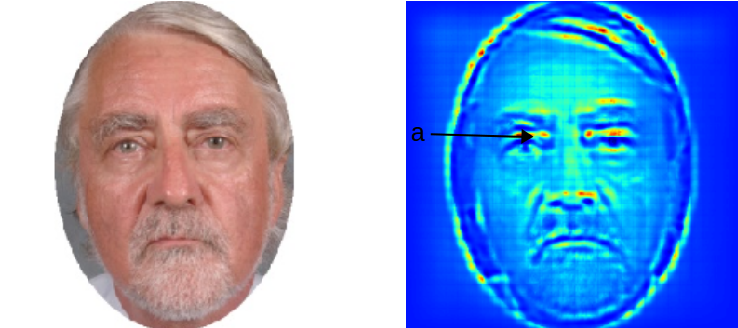}
\caption{
Two old test subjects and the corresponding heatmaps. Indicator arrows are showing areas of high relevance. For the first individual these are the right earlobe (a), the wrinkle below the birthmark (b) and the upper eye bags (c). The second old test-subject has upper eye bags (a) as well as saggy eyelids as the most relevant features that indicated a high age.
}
\label{fig:age}
\end{figure}

The test subjects in figure \ref{fig:age} show three basic features, that according to the model are indicative for advanced age. Indicator arrow (a) in left image shows that the earlobe is seen as evidence for advanced age. Furthermore, indicator arrow (b) points at a significant wrinkle around the eye region. Wrinkles around the eye are indeed indicators of advanced age. However, the most dominant feature indicating advanced age are saggy eyelids.

Gray hair is partially identified as relevant, but not always as the most relevant indicator. Wrinkles, for example, are a more reliable indicator of age.

\subsection{Explanation for Attractiveness}
\begin{figure}[t]
\centering
\includegraphics[width=6cm]{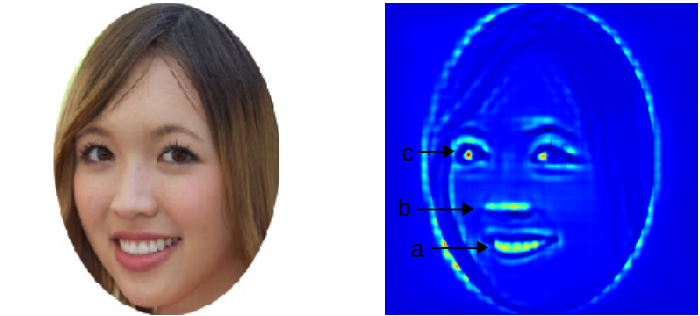}
\includegraphics[width=6cm]{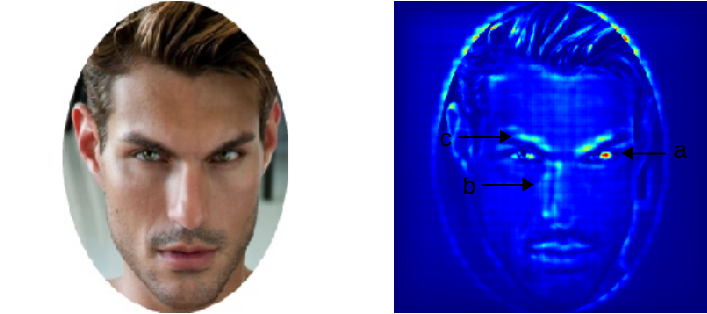}
\caption{
The test subjects have been classified as attractive. For the woman the important features are the eyes (c), the nose (b) and the teeth (a); for the man the hair and the shapes of the eyebrows and the nose.
}
\label{fig:attr}
\end{figure}

Figure \ref{fig:attr} shows two individuals that have been classified as attractive. In both cases, a lot of relevance is assigned to the eyes. Furthermore, straight lines appear in both heatmaps. A possible interpretation is that these are indicators of symmetry. The heatmap of the female shows a straight line below the nose, indicating symmetry of this part. Furthermore, the female person possesses a large space between the eyes and the eyebrows that is considered relevant.

The straight lines in the male example in figure \ref{fig:attr} correspond to the nose, the eyes and partially to the mouth. Asymmetric deviation of the mouth appears to be less relevant. The vertical texture of the man's hair is also assessed as relevant.

\subsection{Occlusion Tests}

\begin{figure}[!htb]
\centering \small
\parbox{1.5cm}{~}%
\parbox{2.25cm}{\centering Original}%
\parbox{2.25cm}{\centering Mouth\\occluded}%
\parbox{2.25cm}{\centering Right eye\\occluded}%
\parbox{2.25cm}{\centering Both eyes\\occluded}\\
\parbox{0.9cm}{~}%
\rotatebox{90}{~~~~~Heatmap~~~~~~~~~~~~~~Heatmap~~~~~~~~~~~~~~~~Image}%
\rotatebox{90}{~(gender-based)~~~~~~~~~(age-based)}%
\includegraphics[width=9cm]{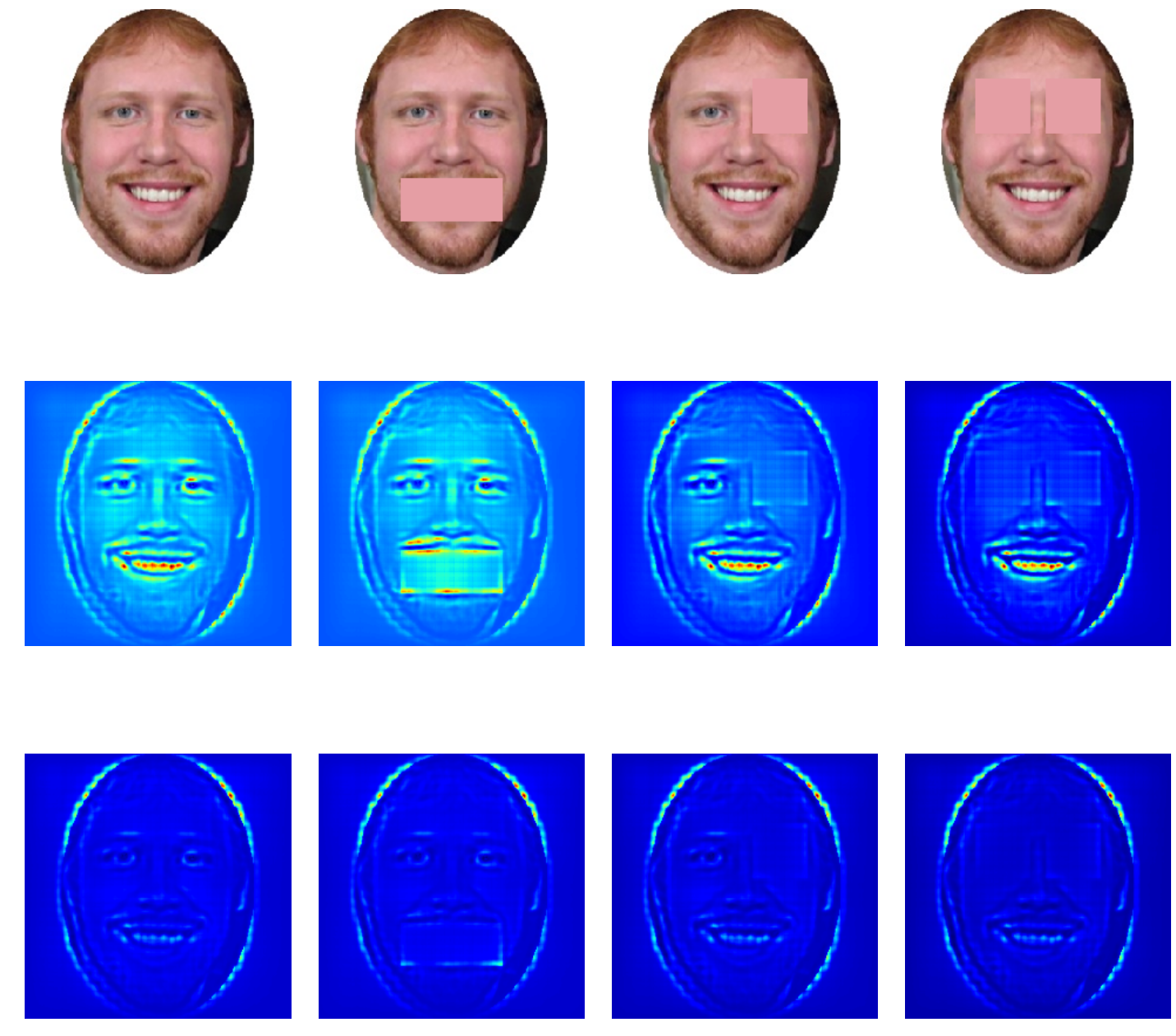}\\
\parbox{5.5cm}{\bf Happiness scores}\parbox{7cm}{~}\\
\parbox{2.5cm}{\it age~based~model}%
\parbox{2.25cm}{\centering 6.69}%
\parbox{2.25cm}{\centering \bf 5.26}%
\parbox{2.25cm}{\centering 6.54}%
\parbox{2.25cm}{\centering 6.73}%
\parbox{1cm}{~}\\
\parbox{2.5cm}{\it gender~based~model}%
\parbox{2.25cm}{\centering 6.62}%
\parbox{2.25cm}{\centering \bf 5.86}%
\parbox{2.25cm}{\centering 6.65}%
\parbox{2.25cm}{\centering 6.75}%
\parbox{1cm}{~}
\caption{Sample subject occluded in various ways shown next to the heatmap and happiness score for the two considered models. Mouth occlusion causes the happiness scores to decrease for both models, and the heatmap values in the mouth region to drop as well.}
\label{fig:occlusion}
\end{figure}

We validate the insights obtained from the analysis of heatmaps by an experiment that analyzes how the model and the heatmaps react to selective occlusions. Figure \ref{fig:occlusion} shows a sample subject where various facial features are occluded: mouth, right eye, and both eyes. We use the skin color for occlusion in order to reduce visual artefacts (e.g. the apparition of strong unnatural edges). Below each image, we show the heatmap and prediction score for two models of happiness: one pretrained on the age model and one pretrained on the gender model.


It can be observed in both cases that only the mouth occlusion is causing the happiness score to decrease significantly. On the other hand, removing the eyes sometimes even increases the perception of happiness, likely due to the attention being ported exclusively to the mouth region.

\section{Conclusion}

Modern learning machines allow the modeling of highly complex and nonlinear problems. Until very recently, it was, however, unclear how to provide explanations for neural network decisions and predictions and what exactly in terms of inputs made the learning machine provide its prediction for each single novel data point. Explanation methods such as \cite{baehrens,ZeilerF14,SimonyanVZ13,DBLP:conf/cidm/LandeckerTBMKB13,BacCVPR16,dosovitskiy2015inverting,zintgraf2016new,yu2014visualizing,nguyen2016multifaceted} have been able to successfully solve this important open problem. The present contribution has applied LRP to a particularly complex domain, namely assessing multifaceted properties in images, such as age, gender, attractivity etc. Humans are excellently trained in their socialization to estimate these properties, perhaps they are less aware of what are the specific features that influence their judgement when distinguishing between an elderly or a younger person. Clearly, wrinkles vs.\ a general percept of freshness would help to holistically guess age. The state-of-the-art convolutive neural nets trained in this work also have a holistic approach to accurately predict age, but the interesting question is, what are the features that are of use for them to come up with this good prediction. Obviously, they differ greatly between young and old faces and it is exactly the relevant features {\em specific per image} that are being extracted by LRP from the convolutive neural net. In addition, we have qualitatively and quantitatively shown their robustness under occlusion. 

In that sense we can probe the artificial neural network on its percept of age, gender, attractivity etc. Future research will aim to relate to psychophysical studies in human perception. After having established
basic convolutive neural network models (with state-of-the-art performance) for predicting certain data rich properties, we could reuse these models and adapt them to data poorer properties such as attractivity by transferring the neural network representation and using only few data point for retraining. While this has been done in a very straight forward manner, it will be a subject of future research to consider how to do this transfer in an optimal manner, also it is interesting to study whether a generative neural network will be better suited to include prior representation knowledge from one prediction task to the other. We would like to stress that the focus of this paper has been to open the black box of a complex learning model that has successfully learned a task where the optimal prediction strategy pursued by the artificial system has been so far somewhat unclear. In particular we are also enabled to judge whether the generalization properties yielding a good out of sample predictor are based on artifacts in a data set or on an appriopriate problem decomposition strategy.     

\section*{Acknowledgement}
This work was supported by the German Ministry for Education and Research as Berlin Big Data Center BBDC (01IS14013A), the Deutsche Forschungsgesellschaft (MU 987/19-1) and the Brain Korea 21 Plus Program through the National Research Foundation of Korea funded by the Ministry of Education. Correspondence to KRM and WS.

\bibliographystyle{splncs03}
\bibliography{egbib}

\begin{thebibliography}{10}
\providecommand{\url}[1]{\texttt{#1}}
\providecommand{\urlprefix}{URL }

\bibitem{ArrACL16}
Arras, L., Horn, F., Montavon, G., M{\"u}ller, K.R., Samek, W.: Explaining
  predictions of non-linear classifiers in nlp. In: Proceedings of the Workshop
  on Representation Learning for NLP at Association for Computational
  Linguistics Conference (ACL) (2016)

\bibitem{bach2015pixel}
Bach, S., Binder, A., Montavon, G., Klauschen, F., M{\"u}ller, K.R., Samek, W.:
  On pixel-wise explanations for non-linear classifier decisions by layer-wise
  relevance propagation. PLOS ONE  10(7),  e0130140 (2015)

\bibitem{baehrens}
Baehrens, D., Schroeter, T., Harmeling, S., Kawanabe, M., Hansen, K.,
  M{\"u}ller, K.R.: How to explain individual classification decisions. The
  Journal of Machine Learning Research  11,  1803--1831 (2010)

\bibitem{bainbridge2013intrinsic}
Bainbridge, W.A., Isola, P., Oliva, A.: The intrinsic memorability of face
  photographs. Journal of Experimental Psychology: General  142(4),  1323
  (2013)

\bibitem{caruana1997multitask}
Caruana, R.: Multitask learning. Machine learning  28(1),  41--75 (1997)

\bibitem{donahue2013decaf}
Donahue, J., Jia, Y., Vinyals, O., Hoffman, J., Zhang, N., Tzeng, E., Darrell,
  T.: Decaf: A deep convolutional activation feature for generic visual
  recognition. arXiv preprint arXiv:1310.1531  (2013)

\bibitem{dosovitskiy2015inverting}
Dosovitskiy, A., Brox, T.: Inverting visual representations with convolutional
  networks. arXiv preprint arXiv:1506.02753  (2015)

\bibitem{goeleven2008karolinska}
Goeleven, E., De~Raedt, R., Leyman, L., Verschuere, B.: The karolinska directed
  emotional faces: a validation study. Cognition and Emotion  22(6),
  1094--1118 (2008)

\bibitem{he2015deep}
He, K., Zhang, X., Ren, S., Sun, J.: Deep residual learning for image
  recognition. arXiv:1512.03385  (2015)

\bibitem{jia2014caffe}
Jia, Y., Shelhamer, E., Donahue, J., Karayev, S., Long, J., Girshick, R.,
  Guadarrama, S., Darrell, T.: Caffe: Convolutional architecture for fast
  feature embedding. In: Proceedings of the ACM International Conference on
  Multimedia. pp. 675--678. ACM (2014)

\bibitem{krizhevsky2012imagenet}
Krizhevsky, A., Sutskever, I., Hinton, G.E.: Imagenet classification with deep
  convolutional neural networks. In: Advances in neural information processing
  systems. pp. 1097--1105 (2012)

\bibitem{DBLP:conf/cidm/LandeckerTBMKB13}
Landecker, W., Thomure, M.D., Bettencourt, L.M.A., Mitchell, M., Kenyon, G.T.,
  Brumby, S.P.: Interpreting individual classifications of hierarchical
  networks. In: {IEEE} Symposium on Computational Intelligence and Data Mining
  ({CIDM}). pp. 32--38 (2013)

\bibitem{BacCVPR16}
Lapuschkin, S., Binder, A., Montavon, G., M{\"u}ller, K.R., Samek, W.:
  Analyzing classifiers: Fisher vectors and deep neural networks. In:
  Proceedings of the IEEE Conference on Computer Vision and Pattern Recognition
  (CVPR) (2016)

\bibitem{LapJMLR16}
Lapuschkin, S., Binder, A., Montavon, G., M{\"u}ller, K.R., Samek, W.: The
  layer-wise relevance propagation toolbox for artificial neural networks. The
  Journal of Machine Learning Research  (2016), in press

\bibitem{le1990handwritten}
Le~Cun, B.B., Denker, J.S., Henderson, D., Howard, R.E., Hubbard, W., Jackel,
  L.D.: Handwritten digit recognition with a back-propagation network. In:
  Advances in neural information processing systems (1990)

\bibitem{lecun2012efficient}
LeCun, Y.A., Bottou, L., Orr, G.B., M{\"u}ller, K.R.: Efficient backprop. In:
  Neural networks: Tricks of the trade, pp. 9--48. Springer (2012)

\bibitem{levi2015age}
Levi, G., Hassner, T.: Age and gender classification using convolutional neural
  networks. In: Proceedings of the IEEE Conference on Computer Vision and
  Pattern Recognition Workshops. pp. 34--42 (2015)

\bibitem{MonArXiv15}
Montavon, G., Bach, S., Binder, A., Samek, W., M{\"u}ller, K.R.: Explaining
  nonlinear classification decisions with deep taylor decomposition.
  arXiv:1512.02479  (2015), \url{http://www.arxiv.org/abs/1509.06321}

\bibitem{nguyen2016multifaceted}
Nguyen, A., Yosinski, J., Clune, J.: Multifaceted feature visualization:
  Uncovering the different types of features learned by each neuron in deep
  neural networks. arXiv preprint arXiv:1602.03616  (2016)

\bibitem{pan2010survey}
Pan, S.J., Yang, Q.: A survey on transfer learning. Knowledge and Data
  Engineering, IEEE Transactions on  22(10),  1345--1359 (2010)

\bibitem{samek2015evaluating}
Samek, W., Binder, A., Montavon, G., Bach, S., M{\"u}ller, K.R.: Evaluating the
  visualization of what a deep neural network has learned. arXiv:1509.06321
  (2015), \url{http://www.arxiv.org/abs/1509.06321}

\bibitem{SimonyanVZ13}
Simonyan, K., Vedaldi, A., Zisserman, A.: Deep inside convolutional networks:
  Visualising image classification models and saliency maps. In: ICLR Workshop
  (2014)

\bibitem{StuArXiv16}
Sturm, I., Bach, S., Samek, W., M{\"u}ller, K.R.: Interpretable deep neural
  networks for single-trial eeg classification. arXiv:1604.08201  (2016),
  \url{http://arxiv.org/abs/1604.08201}

\bibitem{yosinski2014transferable}
Yosinski, J., Clune, J., Bengio, Y., Lipson, H.: How transferable are features
  in deep neural networks? In: Advances in Neural Information Processing
  Systems. pp. 3320--3328 (2014)

\bibitem{yu2014visualizing}
Yu, W., Yang, K., Bai, Y., Yao, H., Rui, Y.: Visualizing and comparing
  convolutional neural networks. arXiv preprint arXiv:1412.6631  (2014)

\bibitem{ZeilerF14}
Zeiler, M.D., Fergus, R.: Visualizing and understanding convolutional networks.
  In: ECCV. pp. 818--833 (2014)

\bibitem{zintgraf2016new}
Zintgraf, L.M., Cohen, T.S., Welling, M.: A new method to visualize deep neural
  networks. arXiv preprint arXiv:1603.02518  (2016)

\end{thebibliography}

\end{document}